%% file: main.tex
\documentclass{svproc}
\usepackage{graphicx}
\usepackage{multicol}
\usepackage{footmisc}
\usepackage{hyperref}
\usepackage{makecell}
\hypersetup{
    colorlinks=true,
    allcolors=blue,      
    }
\usepackage{xurl}
\usepackage{xcolor}

\title{Real-time Vehicle-to-Vehicle Communication Based Network Cooperative Control System through Distributed Database and Multimodal Perception: Demonstrated in Crossroads}

\author{Xinwen Zhu\inst{1*} \and Zihao Li\inst{2*} \and Yuxuan Jiang\inst{3*} \and Jiazhen Xu\inst{1*}  \and Jie Wang\inst{1} \and Xuyang Bai\inst{1}}

\institute{
    Zhejiang University, \email{\{xinwen.23, jie.20, xuyang.20\}@intl.zju.edu.cn, \\ \{12321223\}@zju.edu.cn}
    \and University of Illinois Urbana-Champaign, \email{zihaoli5@illinois.edu}
    \and University of Michigan, \email{jyuxuan@umich.edu}
}   

\authorrunning{Zhu et al.}
\titlerunning{V2V Based Network Cooperative Control System}

\date{January 2024}

\newcommand{\todo}[1]{{\textsf{\textcolor{orange}{[To do]}}}}

\begin{document}

\maketitle

\def\thefootnote{\arabic{footnote}}
\def\thefootnote{*}\footnotetext{These authors contributed equally to this work, which was their undergrad senior design project. This work was done during their undergraduate at Zhejiang University.}

\begin{abstract}
The autonomous driving industry is rapidly advancing, with Vehicle-to-Vehicle (V2V) communication systems highlighting as a key component of enhanced road safety and traffic efficiency. This paper introduces a novel Real-time \textbf{Vehicle-to-Vehicle Communication Based Network Cooperative Control System (VVCCS)}, designed to revolutionize macro-scope traffic planning and collision avoidance in autonomous driving. Implemented on \textbf{Quanser Car (Qcar)} hardware platform, our system integrates the distributed databases into individual autonomous vehicles and an optional central server. We also developed a comprehensive multi-modal perception system with multi-objective tracking and radar sensing. Through a demonstration within a physical crossroad environment, our system showcases its potential to be applied in congested and complex urban environments. The implementations are open-sourced at \url{https://github.com/Essoz/Distributed-Intersection-Traffic-Coordination-With-Lease}

\end{abstract}

\input{body}

\bibliographystyle{acm}
\bibliography{reference}

\input{appendix}

\end{document}

%% file: body.tex
\section{Introduction}

\begin{figure}[htbp!]
    \centering
	\begin{minipage}{0.55\linewidth}
		\centering
		\includegraphics[width=\linewidth]{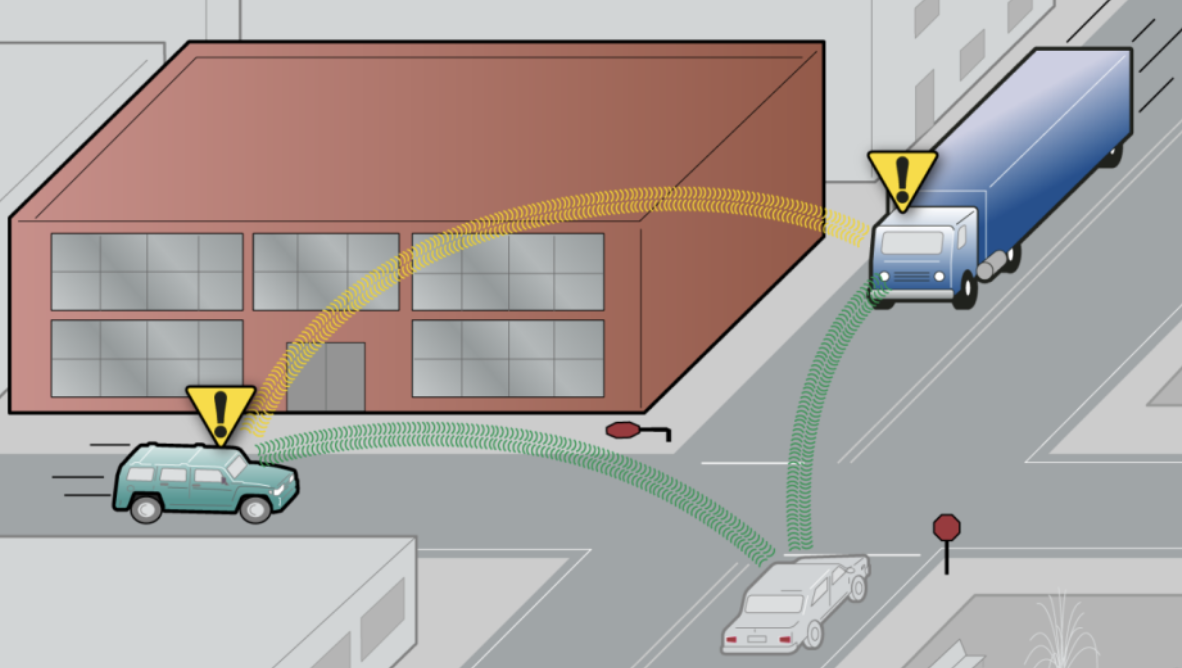}
		\label{intersection}
	\end{minipage}
	\hfill
        \begin{minipage}{0.35\linewidth}
		\centering
		\includegraphics[width=\linewidth]{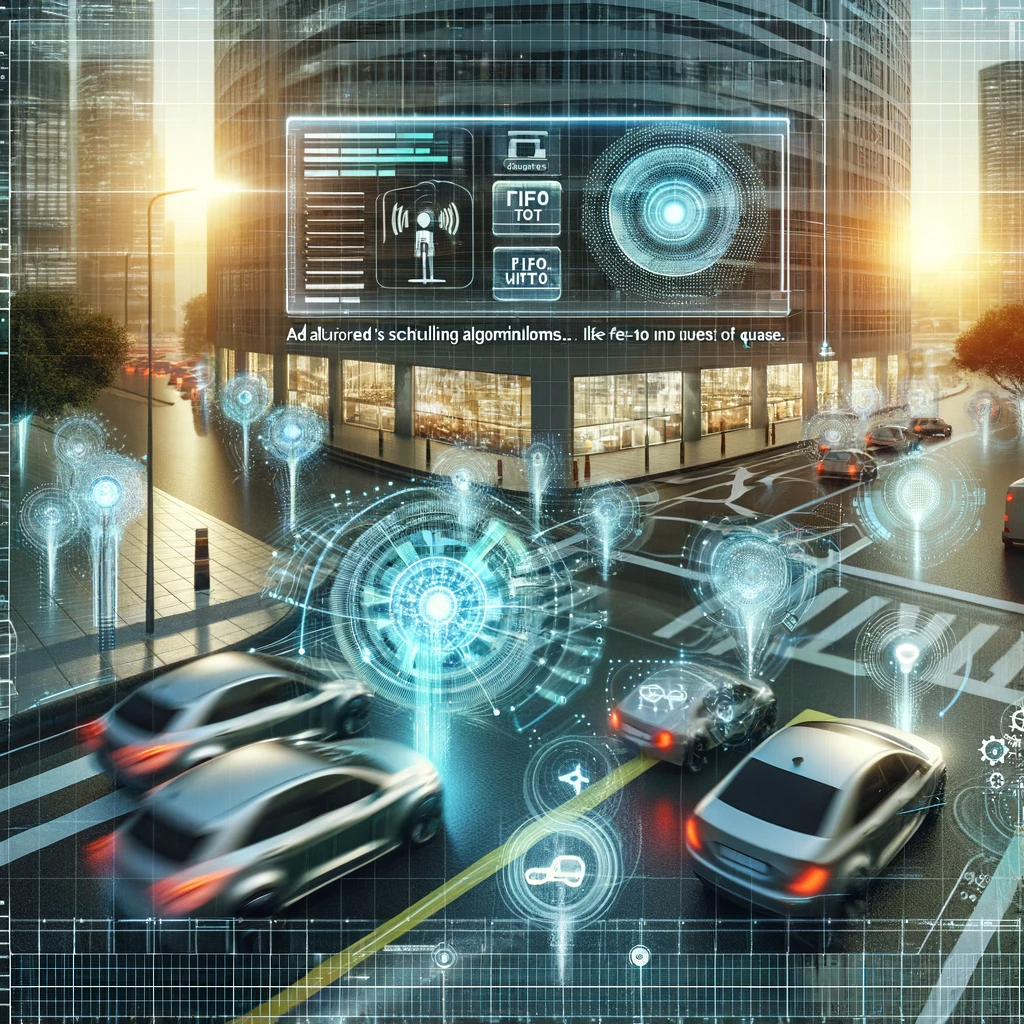}
		\label{sys}
	\end{minipage}
    \caption{Our vision: an advanced V2V communication system in an urban setting, with scheduling algorithms, self-driving, and real-time wireless vehicle interactions \cite{dalle2024v2v}.}
\end{figure}

Crossroads, also known as traffic intersections, represent critical nodes of transportation networks.  However, they also pose a significant risk to transportation safety,erving as hotspots for severe traffic accidents.  Statistically, intersections are disproportionately dangerous: approximately 43\% of all vehicular crashes in the United States occur at or near intersections \cite{LORD2007427}, with similar trends observed globally, such as 40\% of all casualty crashes in Norway and 33\% in Singapore occurring at these junctions \cite{Tay2007Crash}. More alarmingly, these percentages have shown an increasing trend over the years, reflecting a growing concern in urban traffic management.

This situation primarily arises from the following two reasons. First, it’s a common case that vehicles traveling in orthogonal directions cannot notice each other due to the obstruction of buildings. Second, non-motor vehicles and pedestrians are more likely to appear in the blind spot vision of vehicles since there are usually plenty of them gathering in the intersection. These two facts create difficulties for a single vehicle to observe potential collisions and avoid them. Additionally, today, most urban intersections are under passive control mechanisms such as stop signs, which require vehicles to come to a complete stop even when there are no other cars at the intersection. This reduces efficiency by causing unnecessary deceleration. According to a conservative calculation performed by Victor Miller at Stanford University \cite{link}, unnecessary traffic stops in the United States can account for 1.2 billion gallons of consumption per year. Such passive intersection control mechanisms have led to a significant amount of energy waste and call for autonomous and adaptive control mechanisms.

In response to these challenges, this research proposes the V2V-based Network Cooperative Control System (VVCCS), implementing on Quanser Car \cite{qcar} (Qcar) Simulation platform. Aiming at helping the vehicle to get a holistic view of intersection conditions our system make intelligent decisions accordingly. Thereby it enhances safety and operational efficiency V2V autonomous system design while reducing energy consumption.

\section{Related Work}

The development of Autonomous Driving(AD) represents a critical intersection of multiple domain, including communication technologies, autonomous vehicle control, distributed database systems, and multi-modal perception. This section reviews the state-of-the-art research and developments relevant to the components of VVCCS.

\subsection{Vehicular Wireless communication }
According to the survey on vehicular communication by Zeadally et al. \cite{zeadally2020tutorial}, the typical applications based on the V2V can be divided into highly critical and non-critical scenario. VVCCS focused on solving the Pre-crash sensing and Forward collision, which requires effort from both ego car perception and environment infrastructure.

\textbf{Vehicle-to-Server(V2S) Communication:} V2S involve the exchange of data between vehicles and a centralized server. This architecture is instrumental in traffic management, enabling the aggregation and analysis of vehicular data for real-time traffic optimization. Studies by Wang et al. \cite{wang2019v2v} and Qian et al. \cite{Qian2017V2S} have demonstrated V2S's role in reducing congestion and optimizing traffic flow in urban environments. 

\textbf{Vehicle-to-Vehicle(V2V) Communication:} V2V communications facilitate wireless signals to exchange information between vehicles about accidents, weather, roadblocks, and traffic \cite{zeadally2020tutorial}.
This method is crucial for real-time, decentralized decision-making, particularly in situations requiring rapid response, such as collision avoidance \cite{Zhang2022Collision}. Research by Metzner et al. \cite{Metzner2019SituationalAwareness} highlights how V2V can significantly enhance the situational awareness of autonomous vehicles, leading to safer navigation, especially in complex environments like intersections.

\subsection{Distributed Databases in Cooperative Communication System}
The use of distributed databases like Etch Distributed(etcd) \cite{etcd} in AD is increasingly popular due to their high resilience and scalability.  Feng \cite{Feng2023distrib} provided an extensive analysis of distributed database performance in vehicular networks, emphasizing the importance of data consistency and fault tolerance in ensuring reliable communication. Furthermore, the work by Juan et al. \cite{PEREZ2018distrib} showcases the application of distributed databases in forecasting large-scale traffic data, highlighting the advantages of distributed computing in real-time traffic management and analysis.  

\subsection{Multi-modal Perception in Autonomous Vehicles}
Multi-modal perception systems, combining inputs from various sensors such as cameras and LiDAR, are vital in enhancing the situational awareness of autonomous vehicles. Wei et al. \cite{Wei2018fusion} presented a fusion algorithm that integrates camera and LiDAR data for accurate object detection in real time. And Bansal et al. \cite{Bansal2021Risk} classified the collision risk into three level metric. Their method significantly improved the detection accuracy and modeling in complex urban environments. Similarly, Han et al. \cite{Han2018yolo} focused on the improvment of You Only Look Once (YOLO) algorithm \cite{YOLO} in driving sensor fusion, reaching the highly accurate target detection in real time. Further,  Chen et al. \cite{Chen2021LiDARCF} and Hbaieb et al. \cite{Hbaieb2019pedestrian} demonstrated advanced sensor fusion techniques for pedestrian detection within Cooperative Communication System. 

\subsection{Real-time Hardware Platform in Vehicular Technology}
The deployment of real-time systems in vehicular technology is essential for autonomous driving algorithm verification. As demonstrated by Craig et al. \cite{craig2019simu}, simulation is a cornerstone of autonomous driving, allowing testing to occur more rapidly and with significantly less risk than is possible with hardware platforms alone. While Felipe et al. \cite{codevilla2019exploring} argued the simulation methods like Behavior Cloning have many limitation and are hard to generalize into the real-world scenario. 
Retrofitting autonomous driving test platforms like Robosense \cite{robosense}, Baidu Apollo D-kit \cite{baidudkit} seem to be a more realistic approach. However, for V2V communication scenarios specifically, the expensiveness limit their potential to simulate extreme cases like crossroad accident. Quanser Car(Qcar) is a 1/10th scale autonomous vehicle platform designed for research and education in the field of autonomous vehicle technology \cite{qcar}. It is a self-contained system that includes all the necessary hardware and software components to enable autonomous navigation. It stands out as the best balance between safe simulation and algorithm deployment. Its scale and comprehensive integration of necessary technologies make it an ideal platform for experimenting V2V communication protocols and systems in a controlled yet realistic environment. The detailed feature and specification of Qcar is recorded in Appendix A.

In conclusion, the integration of these diverse yet interconnected domains forms the backbone of VVCCS. The ongoing research in these areas continuously contributes to the development of autonmous driving , paving the way for safer, more efficient, and intelligent urban transportation.

\section{System Design: Overview}
\subsection{Subsystem Decomposition}
\begin{figure}[t]
    \centering
    \includegraphics[width=0.8\textwidth]{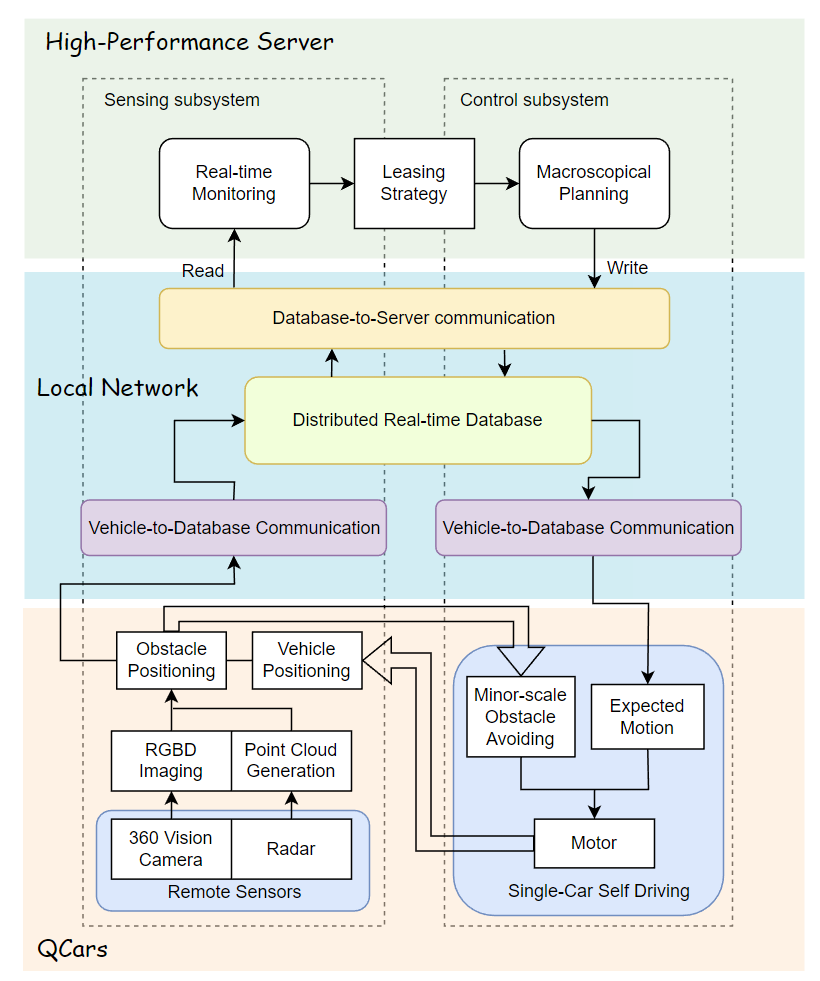}
    \caption{Overview of our Design.}
    \label{sys_final}
\end{figure}

We present the overview of our system in Figure \ref{sys_final}, consisting of four subsystems:
\begin{enumerate}
    \item[1.]\textbf{Control Subsystem}. The control subsystem ensures that the vehicle can run at the desired speed and accelerates/decelerates in time. The vehicle will be controlled by commands with the parameters of pulse-width modulation (PWM) and steering angle. Localization is implemented via the control subsystem by estimating the motor state to infer the position of the vehicle. 
    \item[2.] \textbf{Information Sensing and Multimodal Fusing Subsystem}. We use cameras and the Lidar to sense the environment. Approaching vehicles and pedestrians at the intersection will be recognized and tracked. To realize better precision, information from the camera modality and Lidar modality will be fused before being sent to other vehicles within the V2V network.
    \item[3.]\textbf{Communication Subsystem}. All vehicles equipped with V2V
    communication technology will be able to share their current state (e.g., location, velocity, acceleration, and heading) and their perceptions (locations of detected objects) with each other
    in real time. In the implementation, we set up a distributed database to store the shared information and conduct read-write style communication.
    \item[4.]\textbf{Collision Avoidance}. Respectively, the avoidance algorithm running on the server and 
    vehicles will analyze the data from the shared distributed database and
    single-car recognition subsystems. Global and local decisions will be made on
    the best course of action to avoid potential collisions.
\end{enumerate}

\subsection{Hardware and Software Stacks}
\begin{figure}[htbp!]
    \centering
	\begin{minipage}{0.48\linewidth}
		\centering
		\includegraphics[width=\linewidth]{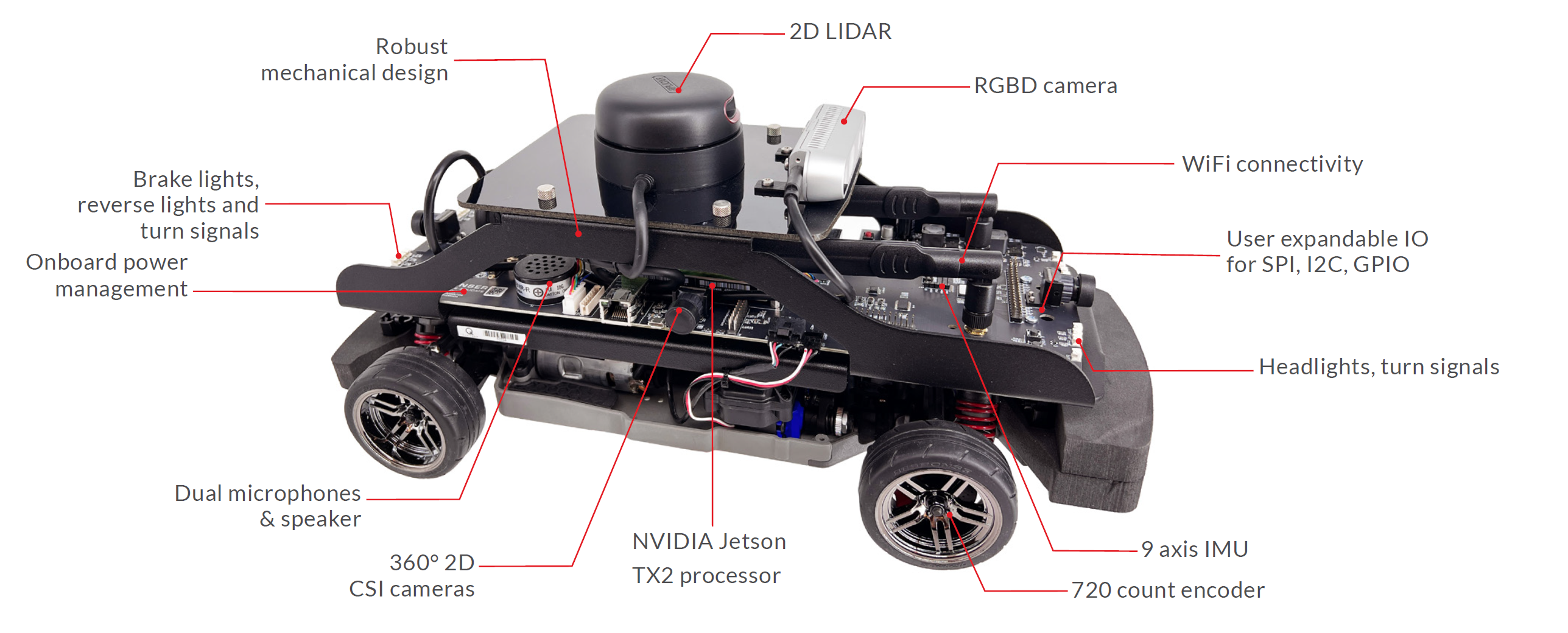}
	\end{minipage}
	\hfill
        \begin{minipage}{0.48\linewidth}
		\centering
		\includegraphics[width=\linewidth]{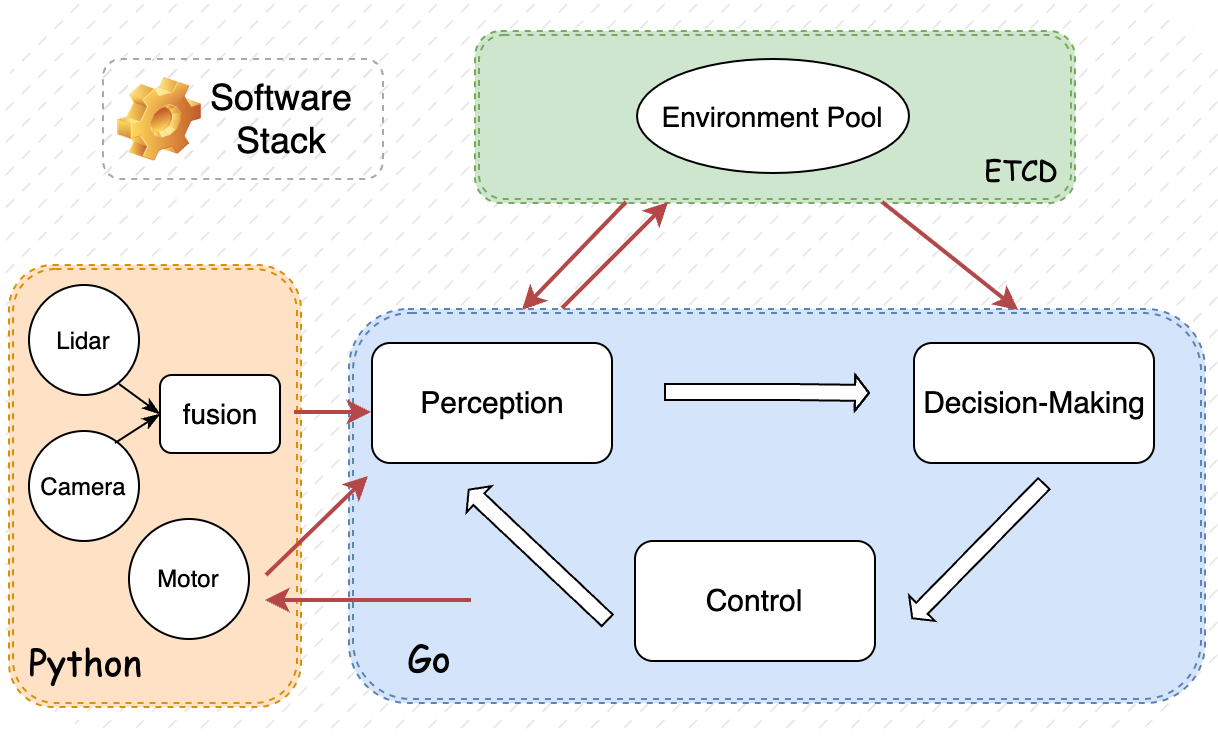}
	\end{minipage}
    \caption{Qcar Diagram and Software Stack}
    \label{phy}
\end{figure}

Our team uses the Quanser Car \cite{qcar} (Qcar) as the experimental car to finish the design. The Qcar (Figure \ref{phy}) is equipped with a LIDAR, an RGBD camera, and four CSI cameras on the front, left, rear, and right sides. Qcar has an on-board GPU to support necessary computation for machine-learning-based multi-objective tracking.

The software stack contains three main blocks (colored in orange, blue, and green). The red arrows between them indicate inter-block data transfers. The orange block refers to Python files that read and process the raw data from the hardware (like Lidar, cameras, and motors). We use Python here because the Qcar itself provides some basic hardware drivers in Python. The green block refers an environment pool implemented in \textit{etcd}, a distributed and reliable database, where we store all observed objects. The blue block refers to the main function of VVCCS. We realize it in Go, since it can both call the Python API and easily interact with etcd. After launching, all three main blocks will be initialized and will run concurrently. 

The main function in Go principally contains a loop with three code blocks: Perception, Decision-Making, and Control. The perception block polls the Python interface to get data. The system will then obtain the state of observed surrounding objects from the Lidar and cameras (after a multimodal information fusion module) and obtain the state of itself from the motor. Thereafter, the perception block will create new objects, delete outdated objects, and update the state of objects that are under tracking in the etcd. The decision-making block runs the whole collision avoidance algorithm with stored data in etcd, and outputs the desired speed to the control block to enforce it. After finishing all three blocks, our system will wait for a certain time before entering the next loop so that it can stay synchronized with the sampling and information fusion module in Python.

\section{System Details}
\subsection{Control subsystem}

This section shows the design and implementation details of the control subsystem of VVCCS. Each physical autonomous driving system requires individual fine-tuning in order to reproduce our experiments. Basically, to apply our system, the following three requirements must be fulfilled: 
\begin{itemize}
    \item The vehicle goes straight when no wheel control command is passed in.
    \item The vehicle can run at a given speed and hold that speed.
    \item The vehicle has the ability to measure the location of itself so that the relative position of different vehicles can be calculated after they receive messages from each other.
\end{itemize}

Minor deviation problems might occur because the diameters of the wheels on the two sides are not perfectly equal, and the suspension may not be perfectly horizontal. As the motor of Qcar is controlled by PWM instead of voltage change, it is needed to build a map from the duty cycle to the actual speed. In our case, we sample and analyze a linear relation of them as shown in Equation \ref{1}
\begin{eqnarray}
		dutyCycle =  k \cdot v_{target}, \  k = 0.1
    \label{1}
\end{eqnarray}

However, the Qcar takes too much time to achieve the desired speed when we want to enforce the collision avoidance algorithm in such a control system. Hence we add feedback terms to realize faster convergence. In Equation \ref{2}, we introduce a proportional error and an integral error so that the system can provide additional power to approach the desired speed.

\begin{eqnarray}
& e(k) = v_{target} - v_{current}
\\
& dutyCycle =  k_{ff} \cdot v_{target}  +  ( k_{p} \cdot e(k) + k_{i}\sum_{i = 0}^{k}e(i)) )
\label{2}
\end{eqnarray}

Lastly, we need to handle the localization issue since Qcar is not equipped with a GPS module. We let the Qcar measure the distance it traveled by monitoring the state of the motor. Once the Qcar can estimate its real-time speed with the state of the motor, we integrate the speed to calculate the distance. Additionally, we hard-code the initial position of Qcars into the system and always run it from there, so that the Qcar can calculate its location accordingly. For an industrial-use autonomous driving car, the process can be simplified by installing GPS.

\subsection{Multimodal Sensing subsystem}

In this section, we will present a detailed description of the sensing subsystem, including its primary components, information processing procedure, and refinement for better performance.

\subsubsection{Primary Components}

\begin{figure}[htbp]
    \centering
	\begin{minipage}{0.40\linewidth}
		\centering
        \includegraphics[width=1\textwidth]{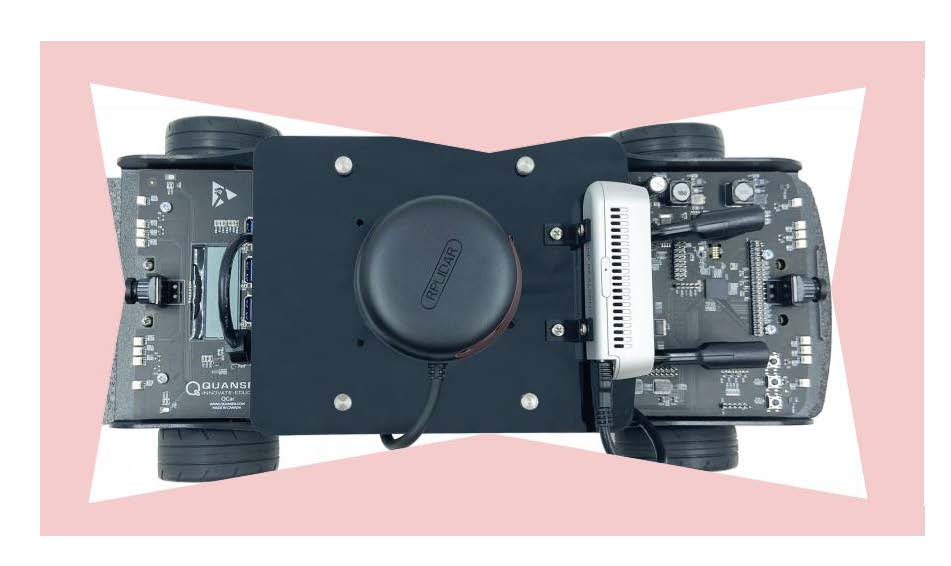}
        \caption{Blind spots of camera}
        \label{blind_spot}
	\end{minipage}
	\hfill
	\begin{minipage}{0.50\linewidth}
		\centering
        \includegraphics[width=1\textwidth]{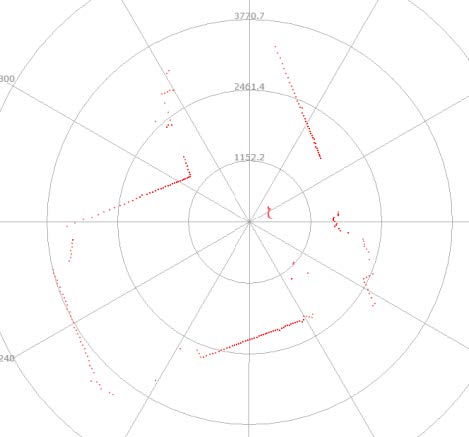}
        \caption{RPLIDAR Demonstration}
        \label{RPLIDAR_mechanism}
	\end{minipage}
\end{figure}

\textbf{Camera.} Qcar contains an Intel D435 RGBD Camera and a 360°2D CSI (Camera Serial Interface) Camera system, as shown in Figure \ref{phy}.  There are four 8MP 2D CSI cameras at the front, left, rear, and right
sides. Each camera has a wide-angle lens providing up to 160°
Horizontal-FOV (field of view) and 120° Vertical-FOV. The blind spots are shown in Figure \ref{blind_spot}.

\noindent\textbf{LIDAR.} The Qcar platform provides RPLIDAR A2M8, an enhanced version of the 2D laser range scanner(LIDAR). It can perform 2D 360-degree scans within a 12-meter range(8-meter range of A2M8-R3 and the below models). It takes up to 8000 samples of laser ranging per second with high rotation speed. The typical scanning frequency of the RPLIDAR A2 is 10hz (600rpm).
Under this condition, the resolution will be 0.45°. During every ranging process, the RPLIDAR emits modulated infrared laser signal and the laser signal is then reflected by the object to be detected. Figure \ref{RPLIDAR_mechanism} provides an example output of RPLIDAR.

\begin{figure}[htbp!]
    \centering
    \includegraphics[width=0.8\linewidth]{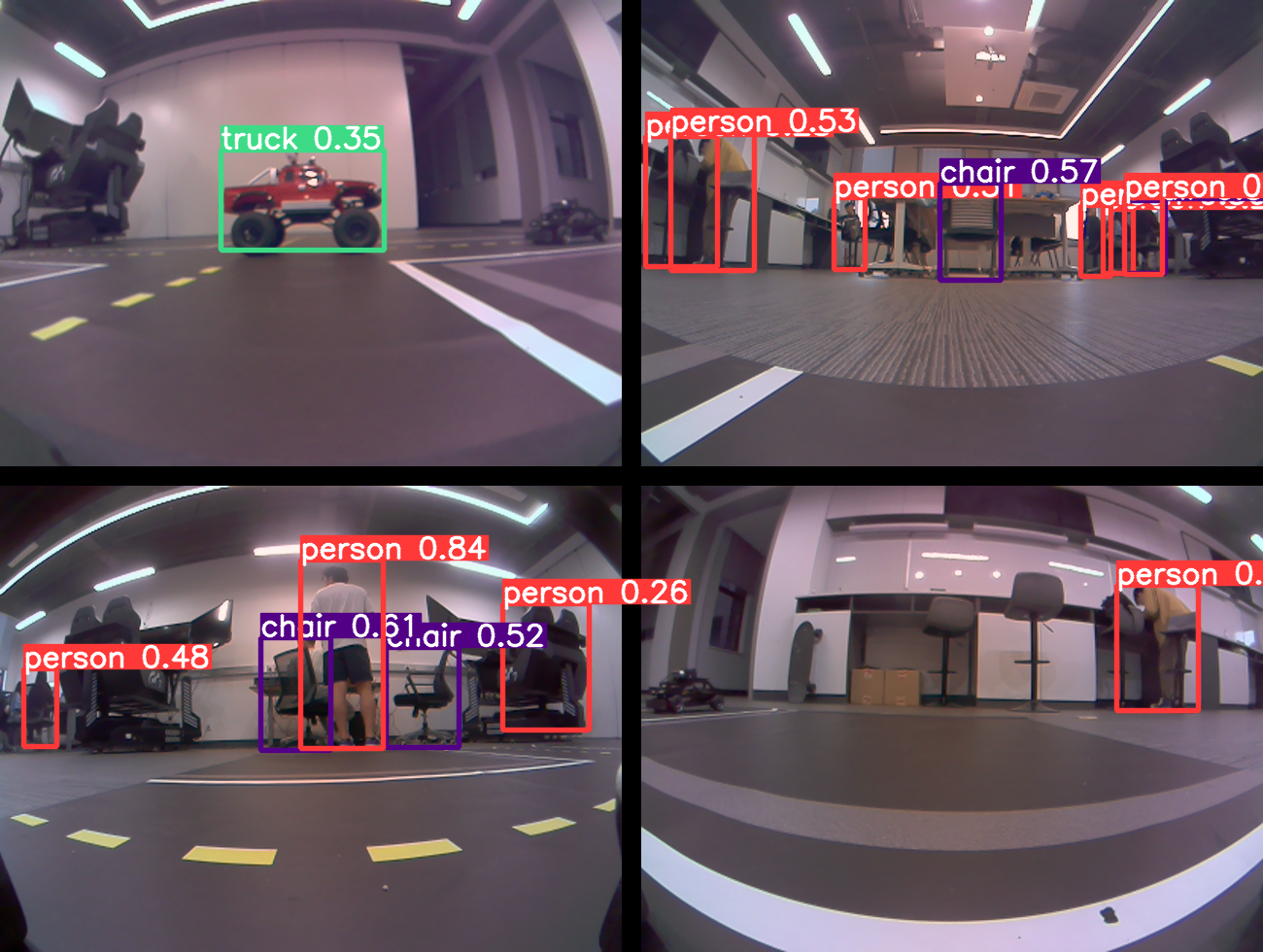}
    \caption{Object Tracking Demonstration}
    \label{object_tracking}
\end{figure}

\noindent\textbf{Object Detection Model.} 
For object detection, the Qcar hardware supports 256 CUDA Core NVIDIA Pascal™ GPU architecture, with 1.3 TFLOPS (FP16) NVIDIA® Jetson™ TX2. We construct one image containing the four initial images from the four cameras as sub-images. We then stream the combined image into the Yolo5 object detection model in real-time. An example figure is shown in Figure \ref{object_tracking}. We tested multiple models and chose the one that has the best performance in our domain. We also tuned the hyper-parameters to 10 frames per second to achieve our real-time requirement. 

\noindent\textbf{Image Coordinates to Angle Conversion.} To get the accurate position of surrounding cars, we need the relationship between the image coordinates of objects and their angle to the Qcar frame. After experiments, we linearly approximate the relationship as $\textbf{angle} = 0.1840*\textbf{pixel} + \textbf{bias}$. We measured the pixel-to-angle correspondence and angle-to-pixel correspondence and concluded that the fitted curve can be approximated linearly on an accurate scale.

\noindent\textbf{Data Fusion.} The angle of surrounding cars to the Qcar can be obtained by processing images from four cameras. In the meantime, LIDAR will provide a point cloud map containing the angle and distance of each point. By extracting points in the range of angles obtained before, we can get type and relative position of the surround objects.

\noindent\textbf{Performance Optimization: Multi-thread Optimization.}
Each loop of our perception system includes four steps: fetching data from each sensor (camera, LIDAR, motor encoder), running the object detection model, fusing prediction from the model with data from the sensors, and sending the command to the control system. However, running such a loop serially will take about $130ms$. 
When fetching data from the sensor, most time is wasted waiting for system API and we mostly use CPU resources. When running an object detection model, GPU resources are mostly used. Before fusing data, prediction from the object detection model must be prepared. So that we can run an object detection model and fuse algorithm in the main loop with some threads for each sensor to fetch their data independently.

\noindent\textbf{Elimination for Duplicates.} In some cases, surrounding objects will be detected by two adjacent cameras at the same time. If they are simply treated as two objects and calculate their relative position separately, the result will be inaccurate. So that we can extract their label box from the prediction list of two adjacent cameras and merge them into one large box. On the other hand, surrounding objects may be detected as two due to the non-max-suppression algorithm of the object detection model. In our code, if two detected objects' relative positions are too close, they will be merged and there will be only one object finally.

\textbf{Data Smoothing: Kalman Filter.} Kalman Filter is an algorithm to compute smoothed time-evolving data given the measured non-smooth data. KF is useful given the noised measured time-varying position.

\subsection{Communication subsystem}

In this section, we present the communication subsystem of VVCCS that supports real-time communication while supporting multi-object tracking.

\noindent\textbf{The etcd Database.} 
etcd \cite{etcd} is a strongly consistent, distributed key-value store that provides a reliable way to store data that needs to be accessed by a distributed system or cluster of machines \cite{DBLP:journals/corr/abs-2409-11585}. etcd exploits the Raft consistency algorithm to coordinate nodes in the cluster. The algorithm elects a master node as leader, who is responsible for synchronization and distribution. When the leader fails, the cluster automatically selects another node as the lead to synchronize the data. Hence, etcd is highly resistant to potential client or communication failure and guarantees the robustness of VVCCS.

\begin{figure}[htbp!]
    \centering
    \includegraphics[width=0.7\linewidth]{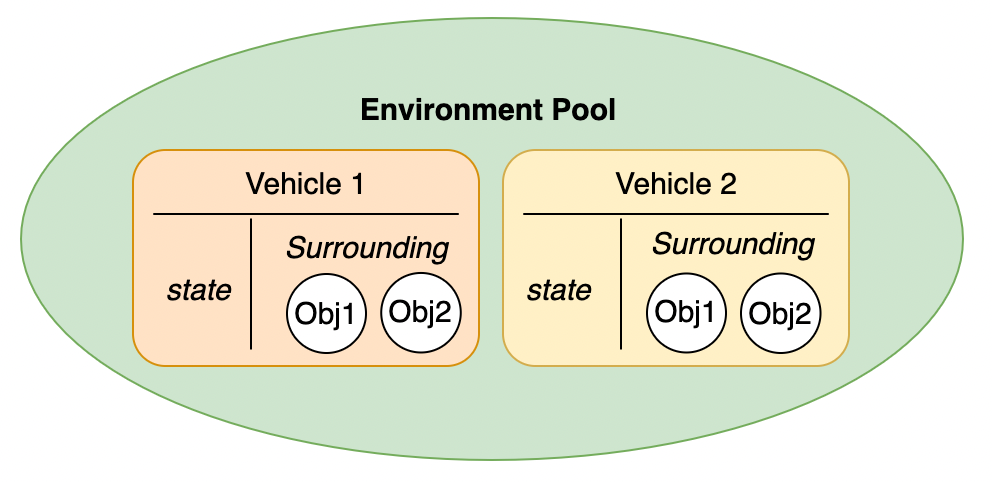}
    \caption{data base structure}
    \label{etcd}
\end{figure}

To implement the communication between vehicles while isolating private data, we construct the following data structure in etcd, as shown in Figure \ref{etcd}. For each vehicle, there are two fields, \textit{state} and \textit{surrounding}, recording the state of the vehicle itself and detected objects, respectively. As etcd provides the functionality of synchronization, the vehicle only needs to update relevant information to its fields. When the vehicle needs to make a decision, it will fetch data from all \textit{surrounding} fields in the etcd, thereby realizing the interaction with other vehicles. An advantage of such implementation is that the communication and planning modules are decoupled.

\textbf{Verification.}  
\textit{Bandwidth} and \textit{latency} are critical criteria for designing the communication module.  We consider a synchronization latency of less than 100ms to be acceptable. In our experiment case, upload and download speed can remain stable at 30MB/s in the WIFI6 local area network. For the communication protocol using method 2, each vehicle updates to etcd by sending a 4KB package every 100ms. A full synchronization among all vehicles takes 2000KB/s, while occasional location updates should take 1/10 of the maximum bandwidth, 200KB/s. Therefore, the estimated bandwidth of our method in the real intersection scenario is approximately 2MB/s, which shows that the laboratory network can support 60 Qcars communicating at the same time. On the other hand, the etcd official benchmark presents that reading one single key after putting has the 90th Percentile Latency of 8.6 ms on 64 clients.

\subsection{Collision Avoidance algorithm} 
\input{collision_avoidance}

\section{Demonstration in Physical Environments}
We set up a physical intersection environment in our lab with a $4.5\times 4.5$ meter canvas depicting the crossroads.
The demonstration videos can be found at \url{https://drive.google.com/drive/folders/1tqCEgUyZGuE_PuScPbvGTvdxDFfrM0GJ?usp=sharing}

\section{Discussion}

\subsection{System Limitation}
Despite the innovative approach of VVCCS, there are inherent limitations that need addressing in future developments. One primary constraint comes from the sensor and CPU computing capabillity of the Qcar platform. The current sensor suite and computation resource, while effective for basic navigation and perception tasks, may not be sufficient for more complex scenarios encountered in real-world driving. Technical advancements are necessary to improve system real-time performance, accuracy, scalability, and reliability. 

\subsection{Future work}
The VVCCS’s current implementation offers a safe and controlled environment for testing emergent driving scenarios without ethical or safety concerns. Future work includes expanding the number and variety of simulations, as well as upgrading the system. One potential direction is to iterate and migrate the system to larger-scale simulation vehicles such as the Baidu Apollo D-kit \cite{baidudkit}, providing a more comprehensive testing ground.

\textbf{Testing and Validation:} Plans for extensive testing and validation in real-world conditions are crucial. This includes trials in diverse environments and weather conditions, along with long-term assessments of operational stability. 

\textbf{Interdisciplinary Integration:} Consideration should be given to integrating the VVCCS with technologies from other fields, such as smart city management and environmental monitoring. This approach could broaden the application scope of the research findings.

\section{Conclusions}
In summary, VVCCS achieves successful collision avoidance in all our experiments at intersections. The overall energy consumption is much lower than that required by traditional traffic control mechanisms, such as traffic lights and stop
signs. In this work, We implement the

precise control of the speed of Qcar by combining the feed-forward and feedback terms in the command to drive the motor and realize the localization by converting the motor state to real speed and conducting the integration. The vision-based object detection subsystem achieves the detection of other
vehicles and pedestrians in the complex environment. Moreover, We exploit the etcd distributed database to make vehicles communicate with each other. Lastly, We propose the idea of the lease to help schedule the timeline for each to pass through the intersection.


%% file: collision_avoidance.tex
This section presents an innovative collision avoidance algorithm. We will delineate the input and output of the algorithm, and expound on how it ensures efficiency and safety in scenarios involving both V2V and non-V2V vehicles.

\noindent\textbf{Algorithm Specification.}
Our algorithm incorporates two subsystems: the lease-based scheduling subsystem and the planning subsystem. The scheduling algorithm employs a series of discrete snapshots capturing the intersection state at given timestamps. Each snapshot records the speed, location, type, and a unique identifier for every traffic participant. From this data, a lease, delineating the temporal duration of a vehicle's intersection occupancy, is created for each vehicle. A lease can be extended, canceled, or reapplied in response to unpredictable circumstances. The planning subsystem evaluates the lease assigned to each vehicle, modulating the vehicle's speed to adhere to the lease terms. In essence, the algorithm processes intersection snapshots as input to produce corresponding vehicle movements as output.

\noindent\textbf{Lease-based Scheduling Subsystem.}
This subsection provides a comprehensive understanding of the elegance, simplicity, and adaptability of our solution. The ingenuity of our design stems from its ability to strike a balance between efficiency, safety, and timeliness depending on specific application scenarios and computational resources. Our design hinges on the proven and effective FIFO (First In First Out) algorithm \cite{coop-driving} used for obstacle avoidance.

\begin{figure}[htb]
    \centering
	\begin{minipage}{0.45\linewidth}
		\centering
		\includegraphics[width=\linewidth]{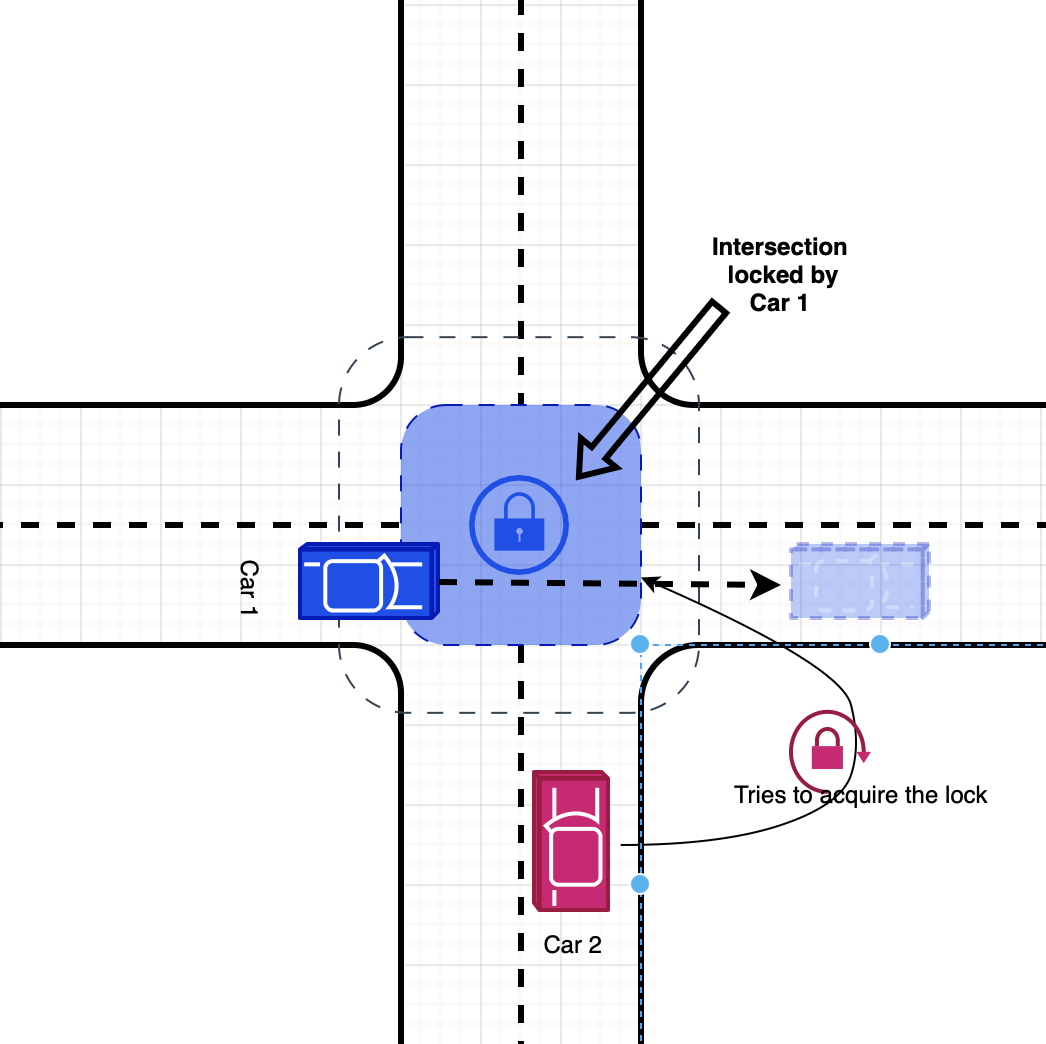}
	\end{minipage}
	\hfill
	\begin{minipage}{0.45\linewidth}
		\centering
		\includegraphics[width=\linewidth]{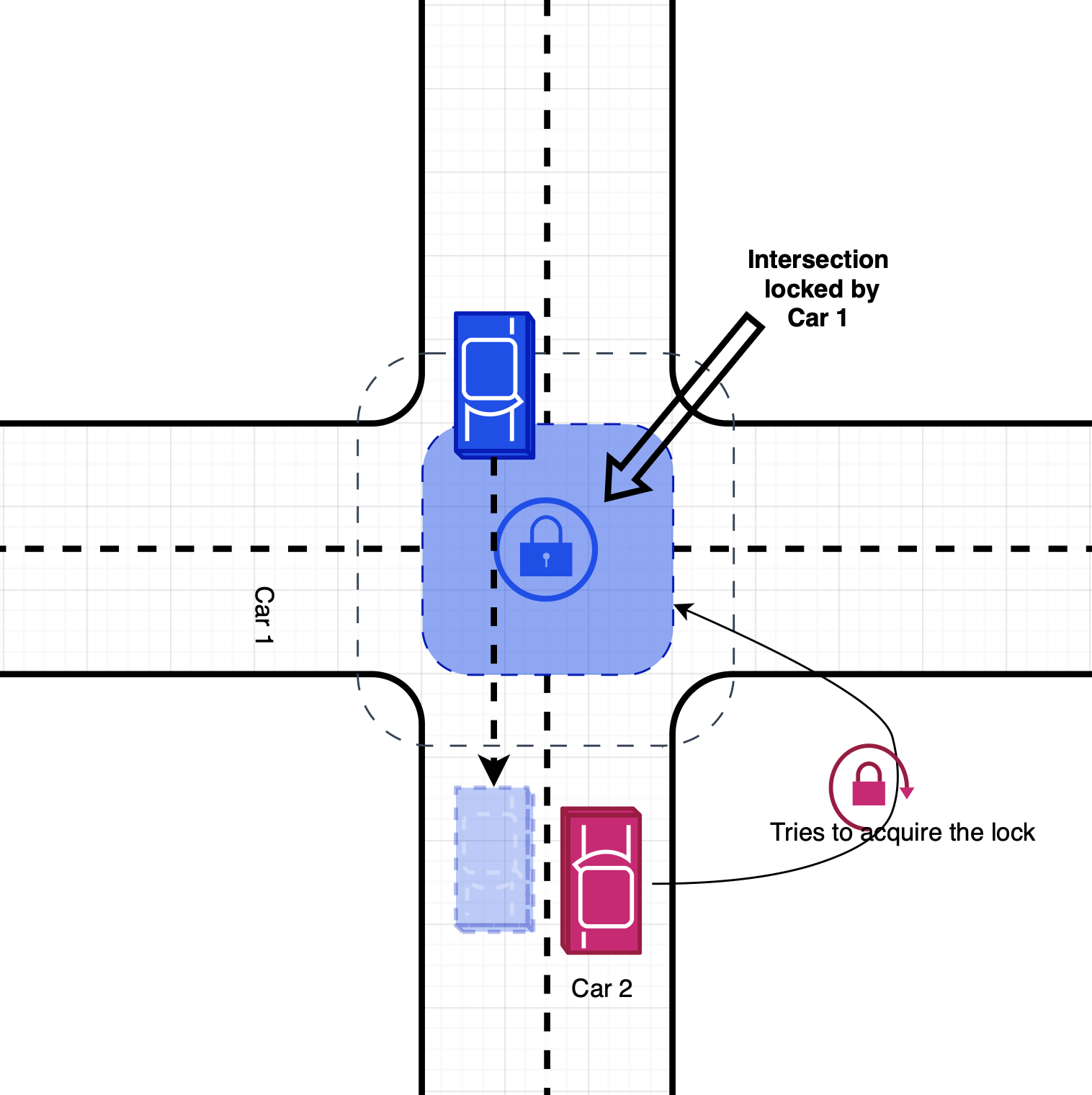}
	\end{minipage}
	\caption{FIFO Algorithm and Low-efficiency of the Lock Implementation}
    \label{fig:fifo}
\end{figure}

\noindent\textbf{The Lock-based Algorithm.}
Previous algorithms metaphorically treat the intersection as a computer science lock. Vehicles attempt to acquire the lock before entering the intersection, delaying their entry if the lock is occupied (Figure \ref{fig:fifo}). This one-at-a-time entry policy assures absolute safety. However, this straightforward approach presents a major drawback: the absence of scheduling capabilities, resulting in safety and efficiency issues. For instance:
\begin{itemize}
    \item \textbf{Abrupt halts:} Vehicles cannot anticipate when a lock might be obtained by others, leading to sudden stops or reduced efficiency.
    \item \textbf{Efficiency conundrums:} Without advance knowledge of lock release, vehicles can't adjust their speed to seamlessly traverse the intersection immediately upon lock availability. This issue exacerbates traffic congestion under heavy traffic conditions.
\end{itemize}

\subsubsection{Our Lease-based Algorithm.}
A lease is akin to a lock, supplemented with a conservative estimation of the duration when a traffic participant is expected to occupy a block. Traffic participants holding currently active leases are permitted to enter the intersections. Safety is guaranteed by ensuring leases do not overlap in time for conflicting paths in the intersection. Our lease-based approach grants a temporal window of safe intersection traversal to each vehicle, allowing them to adjust their speed in anticipation of their assigned lease and improve efficiency.

In a nutshell, every traffic participant's action will be divided into three phases depending on their location.
\begin{itemize}
    \item Planning (before crossing the intersection)
    \item Crossing (inside the intersection)
    \item Post-crossing (after crossing the intersection)
\end{itemize}

\paragraph{The Planning Phase.} In this phase, traffic participants will have two kinds of actions, depending on whether it has made a lease or not. Every traffic participant starts with no lease. To apply for a lease, they must follow these steps:
\begin{itemize}
    \item Estimate the expected time of entering and leaving the intersection area.
    \item Check if there are any conflicting leases.
    \item If there are no conflicting leases, register its lease into etcd, using the expected time.
    \item Else, postpone its lease to the next available slot and register the lease into etcd.
\end{itemize}
These steps guarantee no two leases can overlap during the application phase. After a lease has been acquired, the traffic participant should constantly check the following:
\begin{itemize}
    \item If the current lease can be brought forward? If yes, bring the lease forward to the first available slot. This step is necessary as sometimes, a previous lease can get canceled. In this case, we want to actively check if a lease can be put forward for efficiency concerns.
    \item Check if the current lease is still possible to satisfy within the car's mechanical capabilities. If it is impossible to catch up with a lease anymore or the lease has expired, we want to cancel the lease and reapply the lease.
\end{itemize}

\paragraph{The Crossing Phase.} In this phase, traffic participants mainly do the following for lease management: 
\begin{itemize}
    \item Check if its lease is about to expire. If yes, extend the lease and postpone other participants' leases if necessary, to avoid other participants from entering the intersection before completing leaving.
\end{itemize}

\paragraph{The Post-crossing Phase.} In this phase, traffic participants mainly do the following for lease management: 
\begin{itemize}
    \item Cancel the lease if it is still active. A lease might still be active after the participant has left the intersections because of many factors such as conservative time prediction. We want to cancel the lease early so that other traffic participants can bring their lease forward.
\end{itemize}

\paragraph{Managing Non-V2V Traffic Participants.} Not all traffic participants are equipped with V2V communication capabilities and as such, they may be unable to apply for leases autonomously. In the event of a conflict, we prioritize non-V2V leases by postponing V2V leases instead. This strategy is implemented to minimize the impact of unpredictability from non-V2V participants.

\noindent\textbf{Enforcement}
Once the lease for each vehicle is assigned, the planning subsystem acts as the intermediary between the lease-based scheduling subsystem and the physical layer of vehicle motors, regulating speed and trajectory to meet the scheduling requirements.
\begin{itemize}
    \item \textbf{Planning:} If there is not a lease, the traffic participants should keep going at their current speed. If there is a lease, the participant changes its speed according to the requirement of the lease. If the participant is about to arrive at the intersection but still does not have a lease available, it stops until the leasing system makes a lease.
    \item \textbf{Crossing:} Keep its speed at the advised speed (often set by the government), and stop if the current lease is preempted by a non-V2V traffic participant.
    \item \textbf{Post-crossing:} Keep its speed at the advised speed (often set by the government)
\end{itemize}

The control subsystem and the leasing algorithm, together, will make intersection collision efficient and safe.

\noindent\textbf{Verification}
To verify the properties of the system, we have designed 4 experiments. One for showing the efficiency of the algorithm, and two for showing the safety of the algorithm. The first one is a comparison experiment, the experiment setup contains two V2V vehicles trying to cross the intersection at the same time from different directions and compare the total time for both cars to cross the intersection under our lease-based scheduling algorithm and the lock-based algorithm. We show that our algorithm is consistently 30\% faster. The second one consists of two V2V vehicles trying to cross the intersection at the same time. We show that the lease-based scheduling system can let both cars cross without crashing into each other. The third one consists of two V2V vehicles and one non-V2V vehicle trying to cross the intersection at the same time. We show that without V2V communication, the non-V2V vehicle will crash into the other V2V vehicle as due to visual obstacles, the two cars cannot see each other. However, with our algorithm and communication, the V2V vehicle can slow down to avoid collisions with the non-V2V vehicle. The fourth one consists of two V2V vehicles trying to cross the intersection at the same time. But before they cross, a sudden obstacle arises that blocks their way. We show that the vehicles can do emergency stops and can recover from expired leases right after the obstacle is cleared out of their way. During our tests of the system, we found that the above four experiments have a 100\% success rate on all the experimental verifications.

\noindent\textbf{Engineering Feasibility and Future Improvements.}
In this project, we have shown that the "lease" concept can have great potential in advanced intersection traffic scheduling with a minimum working example. Simple as it is, we want to show that the "lease" concept actually enables further space utilization optimization. For example, we can split the intersection into multiple blocks which have independent lease management systems, to increase the space utilization. Since lease application and the collision avoidance algorithm are largely limited by the accuracy of the prediction algorithm and the movement planning algorithm, the engineering team can easily, based on their specific needs, swap the existing prediction and movement planning algorithms with better ones or simpler ones to balance between performance and amount of computing resource available.

\begin{figure}[!htb]
    \centering
    \includegraphics[scale=0.25]{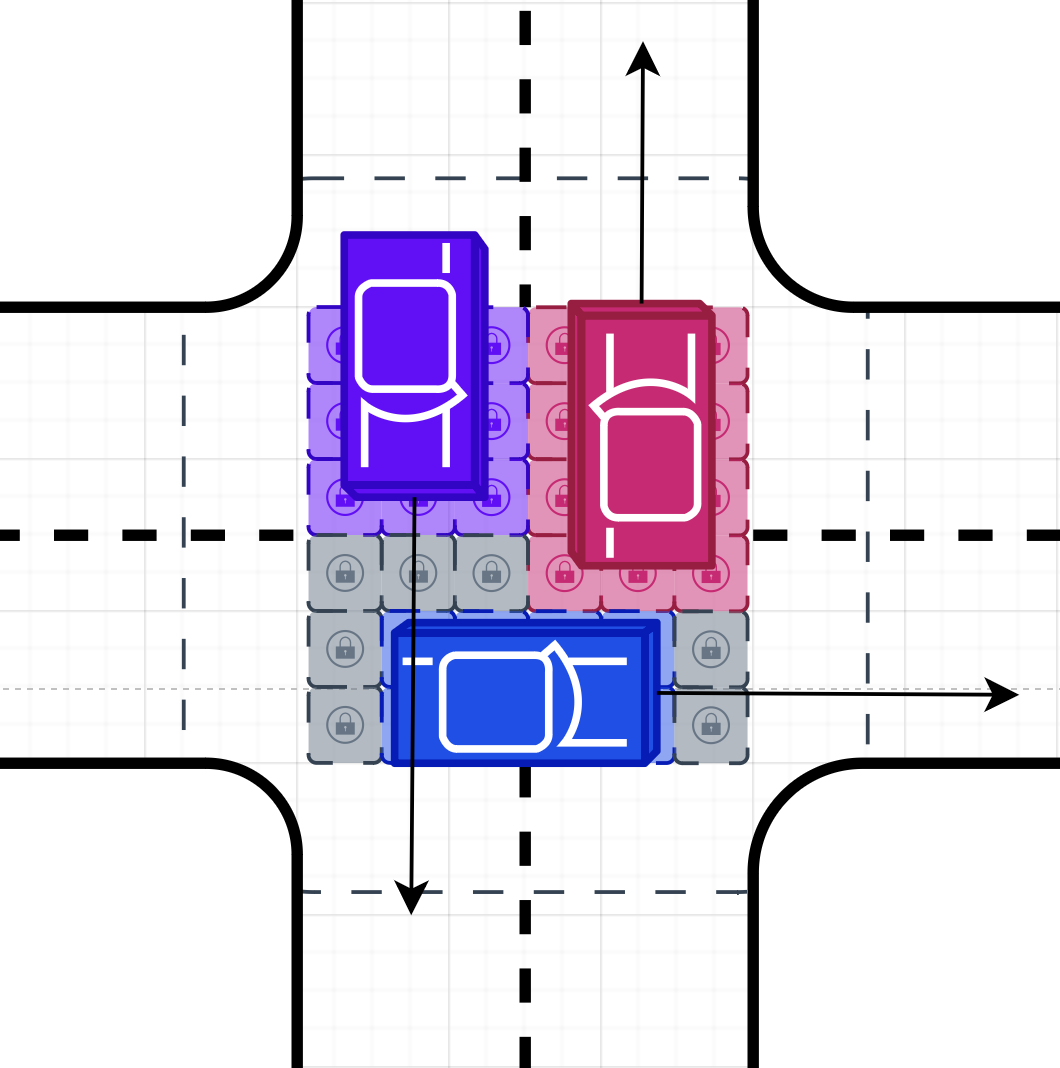}
    \caption{Design of an Improved Lease Algorithm}
    \label{fig:lease_improved}
    \vspace{-5mm}
\end{figure}

%% file: appendix.tex
\newpage

\section{Appendix A: structure of Qcar}

\begin{figure}[htbp]
    \centering
	\begin{minipage}{0.45\linewidth}
		\centering
		\includegraphics[width=\linewidth]{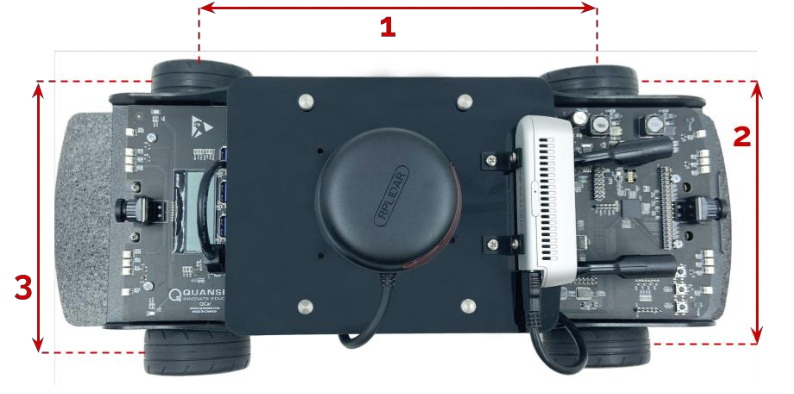}
		\caption{Qcar Dimensions}
		\label{car}
	\end{minipage}
	\hfill
	\begin{minipage}{0.45\linewidth}
		\centering
        \begin{tabular}{|l|l|}
        \hline
        Item   & Value     \\ \hline
        weight & 2.7kg     \\ \hline
        Length & 0.425 m   \\ \hline
        Height & 0.182 m   \\ \hline
        Width  & 0.192 m   \\ \hline
        Tire diameter & 0.066 m \\ \hline
        Wheelbase (Figure \ref{car} \#1)& 0.256 m   \\ \hline
        \makecell[l]{Front and Rear Track  \\ (Figure \ref{car} \#2, 3)} & 0.170 m   \\ \hline
        Maximum steering angle  & ±30° \\ \hline
        \end{tabular}
        \caption{Dimensions}
        \label{Dimensions}
	\end{minipage}
\end{figure}

The dimension of Qcar is shown in Figure \ref{car} and Figure \ref{Dimensions}.

%% file: main.bbl
\begin{thebibliography}{10}

\bibitem{qcar}
Qcar.
\newblock \url{https://www.quanser.com/products/qcar/}.

\bibitem{baidudkit}
{\sc Apollo, B.}
\newblock Robosense lidar systems for autonomous driving.
\newblock \url{https://apollo.baidu.com/community/apollo_d_kit}.
\newblock Accessed: 01.09.2024.

\bibitem{Bansal2021Risk}
{\sc Bansal, A., Singh, J., Verucchi, M., Caccamo, M., and Sha, L.}
\newblock Risk ranked recall: Collision safety metric for object detection systems in autonomous vehicles.
\newblock In {\em 2021 10th Mediterranean Conference on Embedded Computing (MECO)\/} (2021), pp.~1--4.

\bibitem{craig2019simu}
{\sc Brogle, C., Zhang, C., Lim, K.~L., and Bräunl, T.}
\newblock Hardware-in-the-loop autonomous driving simulation without real-time constraints.
\newblock {\em IEEE Transactions on Intelligent Vehicles 4}, 3 (2019), 375--384.

\bibitem{Chen2021LiDARCF}
{\sc Chen, J., Yu, S. C.~H., Tabish, R., Bansal, A., Liu, S., Abdelzaher, T.~F., and Sha, L.~R.}
\newblock Lidar cluster first and camera inference later: A new perspective towards autonomous driving.
\newblock {\em ArXiv abs/2111.09799\/} (2021).

\bibitem{codevilla2019exploring}
{\sc Codevilla, F., Santana, E., L{\'o}pez, A.~M., and Gaidon, A.}
\newblock Exploring the limitations of behavior cloning for autonomous driving.
\newblock In {\em Proceedings of the IEEE/CVF International Conference on Computer Vision\/} (2019), pp.~9329--9338.

\bibitem{etcd}
{\sc etcd company}.
\newblock etcd.
\newblock \url{https://etcd.io/}, 2023.

\bibitem{Feng2023distrib}
{\sc Feng, C., Xu, Z., Zhu, X., Klaine, P.~V., and Zhang, L.}
\newblock Wireless distributed consensus in vehicle to vehicle networks for autonomous driving.
\newblock {\em IEEE Transactions on Vehicular Technology 72}, 6 (2023), 8061--8073.

\bibitem{Han2018yolo}
{\sc Han, J., Liao, Y., Zhang, J., Wang, S., and Li, S.}
\newblock Target fusion detection of lidar and camera based on the improved yolo algorithm.
\newblock {\em Mathematics 6}, 10 (2018).

\bibitem{Hbaieb2019pedestrian}
{\sc Hbaieb, A., Rezgui, J., and Chaari, L.}
\newblock Pedestrian detection for autonomous driving within cooperative communication system.
\newblock In {\em 2019 IEEE Wireless Communications and Networking Conference (WCNC)\/} (2019), pp.~1--6.

\bibitem{dalle2024v2v}
{\sc {Jie Wang, OpenAI DALL-E}}.
\newblock High-tech v2v communication scene.
\newblock \url{https://openai.com/dall-e}, 2024.

\bibitem{coop-driving}
{\sc Li, L., and Wang, F.-Y.}
\newblock Cooperative driving at blind crossings using intervehicle communication.
\newblock {\em IEEE Transactions on Vehicular Technology 55}, 6 (2006), 1712--1724.

\bibitem{DBLP:journals/corr/abs-2409-11585}
{\sc Li, Z., He, S., Yang, Z., Ryu, M., Kim, K., and Madduri, R.~K.}
\newblock Advances in {APPFL:} {A} comprehensive and extensible federated learning framework.
\newblock {\em CoRR abs/2409.11585\/} (2024).

\bibitem{LORD2007427}
{\sc Lord, D., {van Schalkwyk}, I., Chrysler, S., and Staplin, L.}
\newblock A strategy to reduce older driver injuries at intersections using more accommodating roundabout design practices.
\newblock {\em Accident Analysis \& Prevention 39}, 3 (2007), 427--432.

\bibitem{Metzner2019SituationalAwareness}
{\sc Metzner, A., and Wickramarathne, T.}
\newblock Exploiting vehicle-to-vehicle communications for enhanced situational awareness.
\newblock In {\em 2019 IEEE Conference on Cognitive and Computational Aspects of Situation Management (CogSIMA)\/} (2019), pp.~88--92.

\bibitem{link}
{\sc Miller, V.}
\newblock The impact of stopping on fuel consumption.
\newblock \url{http://large.stanford.edu/courses/2011/ph240/miller1/}, 2011.

\bibitem{PEREZ2018distrib}
{\sc Pérez, J.~L., Gutierrez-Torre, A., Berral, J.~L., and Carrera, D.}
\newblock A resilient and distributed near real-time traffic forecasting application for fog computing environments.
\newblock {\em Future Generation Computer Systems 87\/} (2018), 198--212.

\bibitem{Qian2017V2S}
{\sc Qian, L.~P., Wu, Y., Zhou, H., and Shen, X.}
\newblock Dynamic cell association for non-orthogonal multiple-access v2s networks.
\newblock {\em IEEE Journal on Selected Areas in Communications 35}, 10 (2017), 2342--2356.

\bibitem{robosense}
{\sc Robosense}.
\newblock Robosense lidar systems for autonomous driving.
\newblock \url{https://www.robosense.ai/en/scheme}.
\newblock Accessed: 01.09.2024.

\bibitem{Tay2007Crash}
{\sc Tay, R., and Rifaat, S.}
\newblock Factors contributing to the severity of intersection crashes.
\newblock {\em Journal of Advanced Transportation 41\/} (06 2007), 245 -- 265.

\bibitem{YOLO}
{\sc Ultralytics}.
\newblock {YOLO} algorithm documentation, Accessed 2024.
\newblock Online Documentation.

\bibitem{wang2019v2v}
{\sc Wang, R., Xu, Z., Zhao, X., and Hu, J.}
\newblock V2v-based method for the detection of road traffic congestion.
\newblock {\em IET Intelligent Transport Systems 13}, 5 (2019), 880--885.

\bibitem{Wei2018fusion}
{\sc Wei, P., Cagle, L., Reza, T., Ball, J., and Gafford, J.}
\newblock Lidar and camera detection fusion in a real-time industrial multi-sensor collision avoidance system.
\newblock {\em Electronics 7}, 6 (2018).

\bibitem{zeadally2020tutorial}
{\sc Zeadally, S., Guerrero, J., and Contreras, J.}
\newblock A tutorial survey on vehicle-to-vehicle communications.
\newblock {\em Telecommunication Systems 73\/} (March 2020), 469--489.

\bibitem{Zhang2022Collision}
{\sc Zhang, S., Wang, S., Yu, S., Yu, J. J.~Q., and Wen, M.}
\newblock Collision avoidance predictive motion planning based on integrated perception and v2v communication.
\newblock {\em IEEE Transactions on Intelligent Transportation Systems 23}, 7 (2022), 9640--9653.

\end{thebibliography}
